\documentclass[conference]{IEEEtran}

\usepackage{cite}

\ifCLASSINFOpdf
  \usepackage[pdftex]{graphicx}
\else
\fi

\usepackage{amsmath}
\usepackage{amssymb,amsfonts}

\usepackage{algorithm}
\usepackage{algorithmic}

\usepackage{url}

\usepackage{booktabs}
\usepackage{multirow}
\usepackage[table,xcdraw]{xcolor}

\usepackage{orcidlink}

\begin{document}
\title{Classifying Cancer Stage with Open-Source Clinical Large Language Models*\\
{\footnotesize \textsuperscript{*}Note: This manuscript has been accepted to the IEEE International Conference on Healthcare Informatics (IEEE ICHI 2024).}}

% \author{\IEEEauthorblockN{Chia-Hsuan Chang}
% \IEEEauthorblockA{\textit{College of Computing and Informatics} \\
% \textit{Drexel University}\\
% Philadelphia PA, USA \\
% 0000-0001-9116-8244}
% \and
% \IEEEauthorblockN{Mary M. Lucas}
% \IEEEauthorblockA{\textit{College of Computing and Informatics} \\
% \textit{Drexel University}\\
% Philadelphia PA, USA \\
% 0000-0002-0413-7499}
% \and
% \IEEEauthorblockN{Christopher C. Yang}
% \IEEEauthorblockA{\textit{College of Computing and Informatics} \\
% \textit{Drexel University}\\
% Philadelphia PA, USA \\
% 0000-0001-5463-6926}
% \and
% \IEEEauthorblockN{Grace Lu-Yao}
% \IEEEauthorblockA{\textit{Department of Medical Oncology} \\
% \textit{Thomas Jefferson University}\\
% Philadelphia PA, USA \\
% 0000-0002-2925-7737}
% }

\author{\IEEEauthorblockN{Chia-Hsuan Chang\orcidlink{0000-0001-9116-8244}\IEEEauthorrefmark{1},
Mary M. Lucas\orcidlink{0000-0002-0413-7499}\IEEEauthorrefmark{1}, 
Grace Lu-Yao\orcidlink{0000-0002-2925-7737}\IEEEauthorrefmark{2}, and Christopher C. Yang\orcidlink{0000-0001-5463-6926}\IEEEauthorrefmark{1}}
\IEEEauthorblockA{\IEEEauthorrefmark{1}College of Computing and Informatics\\
Drexel University,
Philadelphia PA, USA\\ Email: cc3859,mml367,chris.yang@drexel.edu}
\IEEEauthorblockA{\IEEEauthorrefmark{2}Department of Medical Oncology, Sidney Kimmel Cancer Center\\
Thomas Jefferson University, Philadelphia, PA, USA\\
Email: Grace.LuYao@jefferson.edu}}

\maketitle

\begin{abstract}
Cancer stage classification is important for making treatment and care management plans for oncology patients.  Information on staging is often included in unstructured form in clinical, pathology, radiology and other free-text reports in the electronic health record system, requiring extensive work to parse and obtain. To facilitate the extraction of this information, previous NLP approaches rely on labeled training datasets, which are labor-intensive to prepare. In this study, we demonstrate that without any labeled training data, open-source clinical large language models (LLMs) can extract pathologic tumor-node-metastasis (pTNM) staging information from real-world pathology reports. Our experiments compare LLMs and a BERT-based model fine-tuned using the labeled data. Our findings suggest that while LLMs still exhibit subpar performance in Tumor (T) classification, with the appropriate adoption of prompting strategies, they can achieve comparable performance on Metastasis (M) classification and improved performance on Node (N) classification.

\end{abstract}


\IEEEpeerreviewmaketitle

\section{Introduction}
Although deaths due to cancer have continued to drop in the United States (U.S.), an estimated 2 million people were diagnosed with cancer in 2023 \cite{noauthor_cancer_nodate}. Cancer was one of the leading causes of death in the U.S. in 2021, second only to heart disease, and provisional mortality statistics indicate that this remained unchanged in 2022 and 2023 \cite{noauthor_faststats_2024}. Diagnosing, treating, and monitoring cancer is an interdisciplinary effort that involves multiple health specialties, including medical and surgical oncologists, pathologists, radiologists, interventional radiologists, pharmacists, and nurses among others. All these providers interface with the patient at different times in their medical journey, thereby creating vast amounts of clinical data containing rich clinical insights.  Knowledge of the patient's cancer stage is a critical piece of diagnostic and prognostic information for guiding treatment planning.  An important type of staging data available from pathology reports is the Tumor-Node-Metastasis (TNM) stage, and specifically the pathologic TNM (pTNM) stage (determined after surgery, when the tumor has been excised and tissue samples obtained for analysis \cite{noauthor_cancer_nodate-1}).  TNM is a staging system that allows for a standardized format for presenting information about different cancers. It includes information on the size and extent of the main tumor (T), how much it has spread to the lymph nodes (N), and whether it has spread further to distant sites in the body (M) \cite{noauthor_american_nodate}.  These categories can also be further subdivided to provide additional information. 

While electronic health record (EHR) systems have made it easier to access and analyze large amounts of clinical data for research and for extracting patterns and potential new insights, the ability to parse this kind of staging information at scale remains a challenge.  Clinical data on tumor characteristics and staging are usually contained in clinical free-text notes rather then being recorded in a structured format in the EHR. These notes are often not in a well structured or template format, necessitating the use of natural language processing~(NLP) techniques~\cite{gao_hierarchical_2018,gao_classifying_2019,wu_structured_2020}. While NLP approaches have continued to improve with the advent of pre-trained models~\cite{kefeli_generalizable_2023}, the need for large training datasets remains a challenge~\cite{de_angeli_deep_2021,odisho_natural_2020}. The recent developments in generative large language models (LLMs) provide an opportunity to improve extraction of cancer staging from clinical reports, and to advance cancer research at a more accelerated pace. 

We evaluate the ability of LLMs to extract the pTNM classification from unstructured text data without the need for training datasets. We use pathology reports obtained from The Cancer Genomic Atlas (TCGA) project and compare a general purpose LLM (Llama-2-70b-chat) with clinical LLMs (ClinicalCamel-70B and Med42-70B), assessing their capabilities across different prompting strategies. Evaluating and attempting to improve performance of open-source LLMs on real-world medical data and clinically relevant tasks is important because these LLMs can be installed locally, reducing concerns of exposing protected health information (PHI). 

\section{Literature Review}

NLP has been adopted for extracting information from pathology reports.  Several studies propose specific deep learning models for this purpose. Gao et al. \cite{gao_hierarchical_2018}, and Gao et al. \cite{gao_classifying_2019} proposed a hierarchical network, which learns representation in a layered manner: from words to sentences and reports. They evaluated the generated report representations using pathology reports from the NCI SEER program for five classification tasks, including tumor grade classification. Wu et al. \cite{wu_structured_2020} experimented with the attention-based graph convolution network, in which graph nodes are either words or reports. They defined multiple graphs using different sources of knowledge and applied attention mechanisms to aggregate and propagate the knowledge across graphs, resulting in a better report representation. They used pathology reports from the TCGA project and evaluated various tasks, such as the TNM stage. Rather than introducing a novel neural network, Kelfeli and Tatonetti \cite{kefeli_generalizable_2023} leveraged the power of pre-trained language models. They fine-tuned a clinical-specific model, Clinical-BigBird, for TNM classification using reports from the TCGA project. Their fine-tuned model performed well not only on the testing reports from the TCGA project but also on pathology reports from Columbia University Irving Medical Center. However, all of the above models require substantial amounts of labeled training data, which in turn demands significant amounts of human efforts for curation and annotation.

To reduce the amount of required training reports, Angeli et al.  \cite{de_angeli_deep_2021} investigated active learning techniques for dynamically selecting training samples. They discovered that with the adoption of a convolution neural network, effective active learning techniques can help create a dataset that requires less than half the amount of labeled data to achieve the same performance as a dataset constructed using random sampling. However, it is important to note that even with active learning techniques, a sufficient amount of initial training set and holdout set is still required for a sequence of sample selection processes. Odishi et al.  \cite{odisho_natural_2020} vectorized each token using surrounding contexts, so-called bag-of-n-grams. They learned classifiers for pathologic stage classification using logistic regression, AdaBoost, and random forest. They showed that the model trained using only 64 training reports can generalize well on five times larger testing set of reports. However, the training and testing set were all pathology reports for prostate cancer. The generalizability of their approach across different cancer types is unmeasured.

Recently, LLMs have demonstrated remarkable performance on medical-related tasks owing to their ability to recognize, predict, or generate text or content utilizing transformer models, trained on large volume of publicly available texts. Several studies have indicated that LLMs perform well on various medical Q\&A datasets using very little or without training data~\cite{wu_pmc-llama_2023,han_medalpaca_2023,toma_clinical_2023}. To the best of our knowledge, the performance of LLMs for cancer TNM classification is still unknown. This study aims to fill this gap.

\section{Materials and Methods}

\subsection{Dataset}
The data used for this study comprises free-text pathology reports from the Cancer Genomic Atlas (TCGA) project of the National Cancer Institute (NCI). These reports, in their original format are downloadable as PDF files, with associated metadata found on the NCI Genomic Data Commons (GDC) portal. We utilized a preprocessed corpus that was curated by Kefeli and Tatonetti \cite{kefeli_tcga-reports_2024}. The authors employed optical character recognition (OCR) techniques to convert the PDF reports into machine-readable text and further preprocessed them to remove extraneous information as well as clinically irrelevant headers. This resulted in a dataset of 9,523 reports. A subset of 6,940 reports from this corpus having associated ground truth labels for the T, N, or M stage was then identified for TNM classification task. The reports were split into 85\% training and 15\% testing datasets, using stratified sampling to ensure the class distributions in T, N, and M were similar between datasets. Table~\ref{tab:class_distribution_testing_dataset} presents the class distribution of the testing dataset.

\begin{table}[t]
\caption{Distribution of class labels for T, N, and M data}
\label{tab:class_distribution_testing_dataset}
\centering
\begin{tabular}{@{}llll@{}}
\toprule
           & Labels & Count & Ratio \\ \midrule
\multirow{4}{*}{T14} & 1   & 262 & 0.253 \\
           & 2   & 351 & 0.339 \\
           & 3   & 317 & 0.306 \\
           & 4   & 104 & 0.100 \\ \midrule
\multirow{4}{*}{N03} & 0   & 500 & 0.586 \\
           & 1   & 219 & 0.257 \\
           & 2   & 104 & 0.122 \\
           & 3   & 29  & 0.034 \\ \midrule
\multirow{2}{*}{M01} & 0   & 645 & 0.932 \\
           & 1   & 47  & 0.067 \\ \bottomrule 
\end{tabular}
\end{table}

\subsection{Benchmark}

We employ Clinical-BigBird \cite{li_clinical-longformer_2022}, which extends from the BERT architecture and has 128.1M parameters. Clinical-BigBird outperforms two other well-recognized clinical BERT-based models - BioBERT \cite{lee_biobert_2020} and ClinicalBERT \cite{huang_clinicalbert_2020}, in varying medical-related tasks. To adapt Clinical-BigBird for cancer staging classification task, a previous study \cite{kefeli_generalizable_2023} developed a fine-tuned model using T, N, and M training data split in TCGA dataset and reported its competent performance and generalizability on different sources of pathology reports. Clinical-BigBird is a strong baseline for performance benchmark since it has been exposed to the training split of the TCGA dataset. The model weights of the fine-tuned Clinical-BigBird for TNM staging classification can be downloaded from \url{https://github.com/tatonetti-lab/tnm-stage-classifier}.

\subsection{Large Language Model}

We adopt three open-source LLMs, Llama-2-70b-chat, ClinicalCamel-70B, and Med42-70B, and evaluate their capabilities on cancer pTNM classification.  ClinicalCamel-70B and Med42-70B are clinical LLMs derived from Llama-2 and fine-tuned with clinical data. 

The first model is Llama-2 \cite{touvron_llama_2023}, which has 70B parameters and is pre-trained using two trillion tokens of public texts. Llama-2 has been evaluated in varied different reasoning tasks. We use its dialogue-optimized version, Llama-2-70b-chat, whose model weights can be downloaded from \url{https://ai.meta.com/llama/}, for our experiments. However, Llama-2-70b-chat is a general purpose model. Hence, we survey the clinical LLMs published in 2023 and select ClinicalCamel-70B and Med42-70B as our experimental targets. Both ClinicalCamel-70B and Med42-70B are reported to outperform other open-source models (e.g., ChatDoctor \cite{li_chatdoctor_2023}, MedAlpaca \cite{han_medalpaca_2023}, and PMC-LLAMA \cite{wu_pmc-llama_2023}) and a well-adopted proprietary model, GPT-3.5, in various medical Q\&A datasets.

ClinicalCamel-70B \cite{toma_clinical_2023} is a fine-tune of Llama-2 for clinical research. To adapt the model for accommodating clinical knowledge, the dataset for fine-tuning includes general multi-step conversations, open-access clinical articles, and medical multiple-choice questions with answers. Lastly, Med42-70B\footnote{The details of this model can be found at \url{https://huggingface.co/m42-health/med42-70b}} is also a derived model based on Llama-2, and is fine-tuned using a dataset of 250M tokens compiled from different open-access sources, including medical flashcards, exam questions, and open-domain dialogues. However, unlike the Clinical-BigBird model that we use as the benchmark, these clinical LLMs have not been fine-tuned with the T, N, M training data split from the TCGA dataset. 

We quantize each LLM from float16 to int8 to reduce memory usage load in the loading and inference phases. Our implementation is based on HuggingFace's transformers package, and we run our experiments using two NVIDIA A40 GPUs.

% un-used paragraph:
% To the best of our knowledge, ClinicalCamel-70B and Med42 are two SOTA clinical LLMs~\footnote{We also notice Meditron; however, Meditron is only released the non-instruction-fined model, which is suboptimal in following the instructions.}. We don't compare propriatary LLMs such as Med-PaLM 2 and GPT-4 because the pathology reports contain protected health information (PHI), which may not be uploaded to cloud services. 

\subsection{Prompting strategy}

We implement three different prompting strategies for the three LLMs. The prompting strategy is controlled by different prompt templates~$X$. 

\begin{enumerate}
    \item \textbf{Zero-shot (ZS)} serves as the baseline prompt, in which we provide the relevant context to instruct the model for cancer staging tasks: ``You are provided with a pathology report for a cancer patient. Please review this report and determine the pathologic stage of the patient's cancer.'' The ZS prompt also includes a pathology report. We ask the given LLM to determine the desired TNM staging class.
    \item \textbf{Zero-shot Chain-of-Thoughts (ZS-COT)} \cite{kojima_large_2022} adopts two sequential prompts to perform inference. To better use LLM's capability on reasoning, the first step prompt extends ZS by appending ``Let's think step by step'', triggering the language model to generate step-by-step reasoning for the given report. The second step prompt collects the generated reasoning for the model to perform the TNM staging classification task.
    \item \textbf{Few-shots (FS)} is a widely-adopted prompting strategy with GPT-3.5 \cite{brown_language_2020}. FS extends ZS by providing several text-based demonstrations ($k$-shots) that are relevant to the given task. With FS, the LLM may learn how to solve the given task from demonstrations. In this study, we ask an experienced clinical practitioner to use the TCGA training dataset to select 5-shot, 5-shot, and 6-shot demonstrations for the T, N, and M categories, respectively. All demonstrations are formatted in input-output format, where input is the excerpt extracted from a report, and output is the staging class of the report.  
\end{enumerate}

We adopt greedy decoding when we provide prompts for LLM. Assume the LLM generates a token based on prompt: $p(y|X)$, the greedy decoding in an auto-regressive manner could be represented as:

\begin{equation}
    y_t = \text{argmax}_{y \in Y}p(y|X, y_1, ..., y_{t-1}),
\end{equation}

where $Y$ is the possible token space. After we obtain $\{y_1, ..., y_t\}$ from a model, we use regular expressions (regex) to capture the TNM classification. Specifically, we capture \{T1, T2, T3, and T4\} for T category, \{N0, N1, N2, and N3\} for N category, and \{M0, M1\} for M category. 

\subsection{Performance Metric}

We report the classification performance using precision, recall, and F1 for each stage class (e.g., T1-T4, N0-N3, and M0-M1). All experiments and performance
metrics are based on the test split of the TCGA dataset, to allow for a fair comparison with the benchmark Clinical-BigBird.

\begin{equation}
    \text{precision}= \frac{TP}{TP+FP}
\end{equation}

\begin{equation}
    \text{recall}= \frac{TP}{TP+FN}
\end{equation}

\begin{equation}
    \text{F1}= 2 \times \frac{\text{precision} \times \text{recall}}{\text{precision} + \text{recall}}
\end{equation}

We report macro precision, recall, and F1 for comparing different models and prompting strategies because each stage category has imbalanced class distribution, and the performance of the rare class is equally important as the performance of the frequent class. To deliver robust evaluation, we use bootstrapping resampling to sample each model's predictions and calculate performances $B=500$ times, and each time, we randomly sample $N$ predictions with replacement, where $N$ is the size of the test set. Therefore, we can calculate the 95\% confidence interval for each model's performance metric and perform bootstrapping t-test \cite{efron_introduction_1993}.

\section{Result}

\subsection{Performance Comparison: Benchmark and LLMs with Zero-shot Prompting}

Table~\ref{tab:Performance_T_category}, Table~\ref{tab:Performance_N_category}, and Table~\ref{tab:Performance_M_category} report the performance comparison between Clinical-BigBird (the benchmark) and three different LLMs using zero-shot prompting strategy on T, N, and M staging classification task, respectively. In Table~\ref{tab:Performance_T_category}, we observe that ClinicalCamel-70B and Med42-70B prevail over Llama-2-70b-chat with respect to the macro F1 score. ClinicalCamel-70B has a comparable macro F1 compared with Clinical-BigBird; however, Clinical-BigBird still performs the best in classifying reports in the T category. 

\begin{table}[t]
\caption{Performance table for T category}
\label{tab:Performance_T_category}
\resizebox{\columnwidth}{!}{%
\begin{tabular}{@{}lllll@{}}
\toprule
Model & Class & Precision & Recall & F1-score \\ \midrule
 & T1 & 0.83 & 0.79 & 0.81 \\
 & T2 & 0.76 & 0.84 & 0.80 \\
 & T3 & 0.84 & 0.84 & 0.84 \\
 & T4 & 0.89 & 0.68 & 0.77 \\
\multirow{-5}{*}{Clinical-BigBird} & \cellcolor[HTML]{EFEFEF}Macro avg. & \cellcolor[HTML]{EFEFEF}\begin{tabular}[c]{@{}l@{}}0.83\\ {[}0.80,0.85{]}\end{tabular} & \cellcolor[HTML]{EFEFEF}\begin{tabular}[c]{@{}l@{}}0.79\\ {[}0.76,0.82{]}\end{tabular} & \cellcolor[HTML]{EFEFEF}\begin{tabular}[c]{@{}l@{}}0.81\\ {[}0.78,0.83{]}\end{tabular} \\ \midrule
 & T1 & 0.97 & 0.51 & 0.67 \\
 & T2 & 0.85 & 0.75 & 0.80 \\
 & T3 & 0.56 & 0.96 & 0.70 \\
 & T4 & 0.98 & 0.44 & 0.61 \\
\multirow{-5}{*}{Llama-2-70b-chat + ZS} & \cellcolor[HTML]{EFEFEF}Macro avg. & \cellcolor[HTML]{EFEFEF}\begin{tabular}[c]{@{}l@{}}0.84\\ {[}0.82, 0.86{]}\end{tabular} & \cellcolor[HTML]{EFEFEF}\begin{tabular}[c]{@{}l@{}}0.66\\ {[}0.63,0.70{]}\end{tabular} & \cellcolor[HTML]{EFEFEF}\begin{tabular}[c]{@{}l@{}}0.69\\ {[}0.66,0.73{]}\end{tabular} \\
 & T1 & 0.87 & 0.69 & 0.77 \\
 & T2 & 0.84 & 0.83 & 0.83 \\
 & T3 & 0.73 & 0.88 & 0.80 \\
 & T4 & 0.73 & 0.71 & 0.72 \\
\multirow{-5}{*}{ClinicalCamel-70B + ZS} & \cellcolor[HTML]{EFEFEF}Macro avg. & \cellcolor[HTML]{EFEFEF}\begin{tabular}[c]{@{}l@{}}0.79\\ {[}0.76,0.82{]}\end{tabular} & \cellcolor[HTML]{EFEFEF}\begin{tabular}[c]{@{}l@{}}0.77\\ {[}0.75,0.80{]}\end{tabular} & \cellcolor[HTML]{EFEFEF}\begin{tabular}[c]{@{}l@{}}0.78\\ {[}0.75,0.80{]}\end{tabular} \\
 & T1 & 0.78 & 0.69 & 0.73 \\
 & T2 & 0.93 & 0.70 & 0.80 \\
 & T3 & 0.61 & 0.93 & 0.74 \\
 & T4 & 0.89 & 0.51 & 0.65 \\
\multirow{-5}{*}{Med42-70B + ZS} & \cellcolor[HTML]{EFEFEF}Macro avg. & \cellcolor[HTML]{EFEFEF}\begin{tabular}[c]{@{}l@{}}0.81\\ {[}0.78,0.83{]}\end{tabular} & \cellcolor[HTML]{EFEFEF}\begin{tabular}[c]{@{}l@{}}0.71\\ {[}0.67,0.74{]}\end{tabular} & \cellcolor[HTML]{EFEFEF}\begin{tabular}[c]{@{}l@{}}0.73\\ {[}0.70,0.76{]}\end{tabular} \\ \bottomrule
\end{tabular}%
}
\end{table}

On the other hand, in table~\ref{tab:Performance_N_category}, both ClinicalCamel-70B and Med42-70B achieve over 0.80 macro F1 scores, which not only outperform Llama-2-70b-chat but also Clinical-BigBird. This result suggests that ClinicalCamel-70B and Med42-70B have learned substantial clinical knowledge from their pre-training and fine-tuning stage, releasing the need for labeled training data to perform well on a specific clinical task (i.e., identifying N category from pathology reports).

\begin{table}[t]
\caption{Performance table for N category}
\label{tab:Performance_N_category}
\resizebox{\columnwidth}{!}{%
\begin{tabular}{@{}lllll@{}}
\toprule
Model & Class & Precision & Recall & F1-score \\ \midrule
 & N0 & 0.88 & 0.94 & 0.91 \\
 & N1 & 0.76 & 0.69 & 0.72 \\
 & N2 & 0.73 & 0.52 & 0.61 \\
 & N3 & 0.43 & 0.69 & 0.53 \\
\multirow{-5}{*}{Clinical-BigBird} & \cellcolor[HTML]{EFEFEF}Macro avg. & \cellcolor[HTML]{EFEFEF}\begin{tabular}[c]{@{}l@{}}0.70\\ {[}0.65,0.74{]}\end{tabular} & \cellcolor[HTML]{EFEFEF}\begin{tabular}[c]{@{}l@{}}0.71\\ {[}0.66,0.76{]}\end{tabular} & \cellcolor[HTML]{EFEFEF}\begin{tabular}[c]{@{}l@{}}0.69\\ {[}0.64,0.74{]}\end{tabular} \\ \midrule
 & N0 & 1.00 & 0.09 & 0.16 \\
 & N1 & 0.31 & 0.84 & 0.45 \\
 & N2 & 0.48 & 0.92 & 0.63 \\
 & N3 & 0.75 & 0.54 & 0.63 \\
\multirow{-5}{*}{Llama-2-70b-chat + ZS} & \cellcolor[HTML]{EFEFEF}Macro avg. & \cellcolor[HTML]{EFEFEF}\begin{tabular}[c]{@{}l@{}}0.63\\ {[}0.58,0.68{]}\end{tabular} & \cellcolor[HTML]{EFEFEF}\begin{tabular}[c]{@{}l@{}}0.59\\ {[}0.54,0.64{]}\end{tabular} & \cellcolor[HTML]{EFEFEF}\begin{tabular}[c]{@{}l@{}}0.46\\ {[}0.41,0.51{]}\end{tabular} \\
 & N0 & 0.96 & 0.93 & 0.95 \\
 & N1 & 0.84 & 0.85 & 0.85 \\
 & N2 & 0.73 & 0.84 & 0.78 \\
 & N3 & 0.78 & 0.64 & 0.71 \\
\multirow{-5}{*}{ClinicalCamel-70B + ZS} & \cellcolor[HTML]{EFEFEF}Macro avg. & \cellcolor[HTML]{EFEFEF}\begin{tabular}[c]{@{}l@{}}0.83\\ {[}0.78,0.87{]}\end{tabular} & \cellcolor[HTML]{EFEFEF}\begin{tabular}[c]{@{}l@{}}0.82\\ {[}0.77,0.87{]}\end{tabular} & \cellcolor[HTML]{EFEFEF}\begin{tabular}[c]{@{}l@{}}0.82\\ {[}0.77,0.86{]}\end{tabular} \\
 & N0 & 0.96 & 0.95 & 0.95 \\
 & N1 & 0.84 & 0.84 & 0.84 \\
 & N2 & 0.74 & 0.86 & 0.79 \\
 & N3 & 1.00 & 0.50 & 0.67 \\
\multirow{-5}{*}{Med42-70B + ZS} & \cellcolor[HTML]{EFEFEF}Macro avg. & \cellcolor[HTML]{EFEFEF}\begin{tabular}[c]{@{}l@{}}0.88\\ {[}0.86,0.91{]}\end{tabular} & \cellcolor[HTML]{EFEFEF}\begin{tabular}[c]{@{}l@{}}0.79\\ {[}0.74,0.84{]}\end{tabular} & \cellcolor[HTML]{EFEFEF}\begin{tabular}[c]{@{}l@{}}0.81\\ {[}0.76,0.86{]}\end{tabular} \\ \bottomrule
\end{tabular}%
}
\end{table}

In Table~\ref{tab:Performance_M_category}, Med42-70B performs the best among the other two LLMs, while ClinicalCamel-70B shows a drop in the macro F1 score, performing worse than Llama-2-70b-chat. Clinical-BigBird achieves the highest macro F1 score in M stage classification tasks. When comparing class-specific performance between Clinical-BigBird and Med42-70B, we observe that Clinical-BigBird only has a low recall on M1 (0.19), suggesting it is biased at predicting the negative class. On the contrary, Med42-70B has a higher recall (0.70) while showing a low precision (0.13) in the M1 class. Given that the M1 class represents that cancer has spread to distant parts of the patient's body, the model with a high recall in classifying M1 may help identify patients who need closer monitoring and additional interventions to manage their disease. 

\begin{table}[t]
\caption{Performance table for M category}
\label{tab:Performance_M_category}
\resizebox{\columnwidth}{!}{%
\begin{tabular}{@{}lllll@{}}
\toprule
Model & Class & Precision & Recall & F1-score \\ \midrule
 & M0 & 0.94 & 0.98 & 0.96 \\
 & M1 & 0.47 & 0.19 & 0.27 \\
\multirow{-3}{*}{Clinical-BigBird} & \cellcolor[HTML]{EFEFEF}Macro avg. & \cellcolor[HTML]{EFEFEF}\begin{tabular}[c]{@{}l@{}}0.71\\ {[}0.60,0.82{]}\end{tabular} & \cellcolor[HTML]{EFEFEF}\begin{tabular}[c]{@{}l@{}}0.59\\ {[}0.54,0.64{]}\end{tabular} & \cellcolor[HTML]{EFEFEF}\begin{tabular}[c]{@{}l@{}}0.62\\ {[}0.55,0.68{]}\end{tabular} \\ \midrule
 & M0 & 0.97 & 0.53 & 0.69 \\
 & M1 & 0.11 & 0.80 & 0.19 \\
\multirow{-3}{*}{Llama-2-70b-chat + ZS} & \cellcolor[HTML]{EFEFEF}Macro avg. & \cellcolor[HTML]{EFEFEF}\begin{tabular}[c]{@{}l@{}}0.54\\ {[}0.52,0.56{]}\end{tabular} & \cellcolor[HTML]{EFEFEF}\begin{tabular}[c]{@{}l@{}}0.67\\ {[}0.61,0.73{]}\end{tabular} & \cellcolor[HTML]{EFEFEF}\begin{tabular}[c]{@{}l@{}}0.44\\ {[}0.40,0.48{]}\end{tabular} \\
 & M0 & 0.99 & 0.25 & 0.40 \\
 & M1 & 0.09 & 0.98 & 0.16 \\
\multirow{-3}{*}{ClinicalCamel-70B + ZS} & \cellcolor[HTML]{EFEFEF}Macro avg. & \cellcolor[HTML]{EFEFEF}\begin{tabular}[c]{@{}l@{}}0.54\\ {[}0.53,0.55{]}\end{tabular} & \cellcolor[HTML]{EFEFEF}\begin{tabular}[c]{@{}l@{}}0.61\\ {[}0.58,0.63{]}\end{tabular} & \cellcolor[HTML]{EFEFEF}\begin{tabular}[c]{@{}l@{}}0.28\\ {[}0.25,0.31{]}\end{tabular} \\
 & M0 & 0.97 & 0.66 & 0.79 \\
 & M1 & 0.13 & 0.70 & 0.22 \\
\multirow{-3}{*}{Med42-70B + ZS} & \cellcolor[HTML]{EFEFEF}Macro avg. & \cellcolor[HTML]{EFEFEF}\begin{tabular}[c]{@{}l@{}}0.55\\ {[}0.53,0.57{]}\end{tabular} & \cellcolor[HTML]{EFEFEF}\begin{tabular}[c]{@{}l@{}}0.68\\ {[}0.61,0.75{]}\end{tabular} & \cellcolor[HTML]{EFEFEF}\begin{tabular}[c]{@{}l@{}}0.50\\ {[}0.46,0.54{]}\end{tabular} \\ \bottomrule
\end{tabular}%
}
\end{table}

We also observe an interesting phenomenon: All models exhibit worse macro F1 scores for rare classes (i.e., T4, N3, and M1) than for common classes. The three adopted LLMs have not been exposed to the class distribution but still show performance degradation on rare classes. The reason may be the difficulty in identifying rare classes by nature, but the finding indicates a potential margin for performance improvement.  

\subsection{Performance Comparison: LLMs with Different Prompting Strategies}

Table~\ref{tab:performance-comparison-prompting-strategies} reports the performance comparison of three LLMs using different prompting strategies. We observe that the reasoning generated by the models themselves benefits their classification performance. With ZS-COT, Llama-2-70b-chat substantially improves macro F1 scores on all three categories, reducing the performance gaps between it and the two clinical-specific LLMs. ClinicalCamel-70B with ZS-COT retains the same macro F1 on the T and N categories while improving macro F1 on the M category. ZS-COT also assists Med42-70B in achieving higher macro F1 scores in all three categories. As a result, we conclude that for this task, ZS-COT is a better prompting strategy than ZS.

We observe mixed results when we use FS prompting. In most cases, the three LLMs have worse macro F1 compared with ZS performance. The notable exception is Med42-70B which reached a competent macro F1 score (0.62) on M category compared with Clinical-BigBird. We attribute this performance degradation effect to the fact that pathology reports in the TCGA dataset come from different sources and are therefore written in different styles and formats. Given that LLMs are sensitive to the prompt format \cite{sclar_quantifying_2023}, we point out this observation for further studies.

\begin{table*}[t]
\caption{Performance comparison of models using different prompting strategies}
\label{tab:performance-comparison-prompting-strategies}
\resizebox{\textwidth}{!}{%
\begin{tabular}{@{}lllllllllll@{}}
\toprule
 &  & \multicolumn{3}{l}{Zero-shot} & \multicolumn{3}{l}{ZS-COT} & \multicolumn{3}{l}{Few-shots} \\
 & Model & Macro P & Macro R & Macro F1 & Macro P & Macro R & Macro F1 & Macro P & Macro R & Macro F1 \\ \midrule
\multirow{3}{*}{T Category} & Llama-2-70b-chat & \begin{tabular}[c]{@{}l@{}}0.84\\ {[}0.82, 0.86{]}\end{tabular} & \begin{tabular}[c]{@{}l@{}}0.66\\ {[}0.63,0.70{]}\end{tabular} & \begin{tabular}[c]{@{}l@{}}0.69\\ {[}0.66,0.73{]}\end{tabular} & \begin{tabular}[c]{@{}l@{}}0.82\\ {[}0.79,0.84{]}\end{tabular} & \begin{tabular}[c]{@{}l@{}}0.72\\ {[}0.69,0.75{]}\end{tabular} & \begin{tabular}[c]{@{}l@{}}0.75\\ {[}0.71,0.78{]}\end{tabular} & \begin{tabular}[c]{@{}l@{}}0.74\\ {[}0.71,0.78{]}\end{tabular} & \begin{tabular}[c]{@{}l@{}}0.68\\ {[}0.65,0.71{]}\end{tabular} & \begin{tabular}[c]{@{}l@{}}0.69\\ {[}0.66,0.72{]}\end{tabular} \\
 & ClinicalCamel-70B & \begin{tabular}[c]{@{}l@{}}0.79\\ {[}0.76,0.82{]}\end{tabular} & \begin{tabular}[c]{@{}l@{}}0.77\\ {[}0.75,0.80{]}\end{tabular} & \begin{tabular}[c]{@{}l@{}}0.78\\ {[}0.75,0.80{]}\end{tabular} & \begin{tabular}[c]{@{}l@{}}0.78\\ {[}0.75,0.81{]}\end{tabular} & \begin{tabular}[c]{@{}l@{}}0.78\\ {[}0.76,0.81{]}\end{tabular} & \begin{tabular}[c]{@{}l@{}}0.78\\ {[}0.75,0.81{]}\end{tabular} & \begin{tabular}[c]{@{}l@{}}0.66\\ {[}0.63,0.69{]}\end{tabular} & \begin{tabular}[c]{@{}l@{}}0.66\\ {[}0.62,0.69{]}\end{tabular} & \begin{tabular}[c]{@{}l@{}}0.64\\ {[}0.61,0.67{]}\end{tabular} \\
 & Med42-70B & \begin{tabular}[c]{@{}l@{}}0.81\\ {[}0.78,0.83{]}\end{tabular} & \begin{tabular}[c]{@{}l@{}}0.71\\ {[}0.67,0.74{]}\end{tabular} & \begin{tabular}[c]{@{}l@{}}0.73\\ {[}0.70,0.76{]}\end{tabular} & \begin{tabular}[c]{@{}l@{}}0.80\\ {[}0.77,0.83{]}\end{tabular} & \begin{tabular}[c]{@{}l@{}}0.77\\ {[}0.74,0.80{]}\end{tabular} & \begin{tabular}[c]{@{}l@{}}0.78\\ {[}0.75,0.81{]}\end{tabular} & \begin{tabular}[c]{@{}l@{}}0.74\\ {[}0.71,0.78{]}\end{tabular} & \begin{tabular}[c]{@{}l@{}}0.75\\ {[}0.71,0.78{]}\end{tabular} & \begin{tabular}[c]{@{}l@{}}0.74\\ {[}0.71,0.78{]}\end{tabular} \\ \midrule
\multirow{3}{*}{N Category} & Llama-2-70b-chat & \begin{tabular}[c]{@{}l@{}}0.63\\ {[}0.58,0.68{]}\end{tabular} & \begin{tabular}[c]{@{}l@{}}0.59\\ {[}0.54,0.64{]}\end{tabular} & \begin{tabular}[c]{@{}l@{}}0.46\\ {[}0.41,0.51{]}\end{tabular} & \begin{tabular}[c]{@{}l@{}}0.69\\ {[}0.64,0.74{]}\end{tabular} & \begin{tabular}[c]{@{}l@{}}0.76\\ {[}0.71,0.81{]}\end{tabular} & \begin{tabular}[c]{@{}l@{}}0.70\\ {[}0.65,0.74{]}\end{tabular} & \begin{tabular}[c]{@{}l@{}}0.61\\ {[}0.55,0.67{]}\end{tabular} & \begin{tabular}[c]{@{}l@{}}0.57\\ {[}0.52,0.62{]}\end{tabular} & \begin{tabular}[c]{@{}l@{}}0.47\\ {[}0.42,0.53{]}\end{tabular} \\
 & ClinicalCamel-70B & \begin{tabular}[c]{@{}l@{}}0.83\\ {[}0.78,0.87{]}\end{tabular} & \begin{tabular}[c]{@{}l@{}}0.82\\ {[}0.77,0.87{]}\end{tabular} & \begin{tabular}[c]{@{}l@{}}0.82\\ {[}0.77,0.86{]}\end{tabular} & \begin{tabular}[c]{@{}l@{}}0.84\\ {[}0.79,0.89{]}\end{tabular} & \begin{tabular}[c]{@{}l@{}}0.82\\ {[}0.77,0.87{]}\end{tabular} & \begin{tabular}[c]{@{}l@{}}0.82\\ {[}0.78,0.87{]}\end{tabular} & \begin{tabular}[c]{@{}l@{}}0.78\\ {[}0.75,0.81{]}\end{tabular} & \begin{tabular}[c]{@{}l@{}}0.69\\ {[}0.63,0.76{]}\end{tabular} & \begin{tabular}[c]{@{}l@{}}0.68\\ {[}0.60,0.74{]}\end{tabular} \\
 & Med42-70B & \begin{tabular}[c]{@{}l@{}}0.88\\ {[}0.86,0.91{]}\end{tabular} & \begin{tabular}[c]{@{}l@{}}0.79\\ {[}0.74,0.84{]}\end{tabular} & \begin{tabular}[c]{@{}l@{}}0.81\\ {[}0.76,0.86{]}\end{tabular} & \begin{tabular}[c]{@{}l@{}}0.84\\ {[}0.78,0.89{]}\end{tabular} & \begin{tabular}[c]{@{}l@{}}0.81\\ {[}0.76,0.86{]}\end{tabular} & \begin{tabular}[c]{@{}l@{}}0.82\\ {[}0.77,0.87{]}\end{tabular} & \begin{tabular}[c]{@{}l@{}}0.82\\ {[}0.76,0.88{]}\end{tabular} & \begin{tabular}[c]{@{}l@{}}0.57\\ {[}0.52,0.63{]}\end{tabular} & \begin{tabular}[c]{@{}l@{}}0.64\\ {[}0.58,0.70{]}\end{tabular} \\ \midrule
\multirow{3}{*}{M Category} & Llama-2-70b-chat & \begin{tabular}[c]{@{}l@{}}0.54\\ {[}0.52,0.56{]}\end{tabular} & \begin{tabular}[c]{@{}l@{}}0.67\\ {[}0.61,0.73{]}\end{tabular} & \begin{tabular}[c]{@{}l@{}}0.44\\ {[}0.40,0.48{]}\end{tabular} & \begin{tabular}[c]{@{}l@{}}0.55\\ {[}0.53,0.58{]}\end{tabular} & \begin{tabular}[c]{@{}l@{}}0.68\\ {[}0.61,0.76{]}\end{tabular} & \begin{tabular}[c]{@{}l@{}}0.51\\ {[}0.47,0.55{]}\end{tabular} & \begin{tabular}[c]{@{}l@{}}0.51\\ {[}0.48,0.53{]}\end{tabular} & \begin{tabular}[c]{@{}l@{}}0.51\\ {[}0.46,0.56{]}\end{tabular} & \begin{tabular}[c]{@{}l@{}}0.19\\ {[}0.16,0.22{]}\end{tabular} \\
 & ClinicalCamel-70B & \begin{tabular}[c]{@{}l@{}}0.54\\ {[}0.53,0.55{]}\end{tabular} & \begin{tabular}[c]{@{}l@{}}0.61\\ {[}0.58,0.63{]}\end{tabular} & \begin{tabular}[c]{@{}l@{}}0.28\\ {[}0.25,0.31{]}\end{tabular} & \begin{tabular}[c]{@{}l@{}}0.55\\ {[}0.53,0.58{]}\end{tabular} & \begin{tabular}[c]{@{}l@{}}0.67\\ {[}0.59,0.75{]}\end{tabular} & \begin{tabular}[c]{@{}l@{}}0.54\\ {[}0.49,0.58{]}\end{tabular} & \begin{tabular}[c]{@{}l@{}}0.52\\ {[}0.50,0.54{]}\end{tabular} & \begin{tabular}[c]{@{}l@{}}0.57\\ {[}0.50,0.62{]}\end{tabular} & \begin{tabular}[c]{@{}l@{}}0.33\\ {[}0.29,0.36{]}\end{tabular} \\
 & Med42-70B & \begin{tabular}[c]{@{}l@{}}0.55\\ {[}0.53,0.57{]}\end{tabular} & \begin{tabular}[c]{@{}l@{}}0.68\\ {[}0.61,0.75{]}\end{tabular} & \begin{tabular}[c]{@{}l@{}}0.50\\ {[}0.46,0.54{]}\end{tabular} & \begin{tabular}[c]{@{}l@{}}0.55\\ {[}0.53,0.58{]}\end{tabular} & \begin{tabular}[c]{@{}l@{}}0.68\\ {[}0.61,0.76{]}\end{tabular} & \begin{tabular}[c]{@{}l@{}}0.53\\ {[}0.49,0.57{]}\end{tabular} & \begin{tabular}[c]{@{}l@{}}0.62\\ {[}0.56,0.69{]}\end{tabular} & \begin{tabular}[c]{@{}l@{}}0.62\\ {[}0.56,0.69{]}\end{tabular} & \begin{tabular}[c]{@{}l@{}}0.62\\ {[}0.56,0.68{]}\end{tabular} \\ \bottomrule
\end{tabular}%
}
\end{table*}

Table~\ref{tab:t-statistics} reports the bootstrapping t-test results to compare the difference in macro F1 between Clinical-BigBird and the best model + prompting strategy in each category, selected from Table~\ref{tab:performance-comparison-prompting-strategies}. In the table, we report t-statistics (macro F1 of Clinical-BigBird minus macro F1 of the selected model) and indicate the significance $^{***}$ if p-value $< 0.05$. Our results show that Clinical Big Bird is significantly better than the best LLM, Med42-70B + ZS-COT, in the T category. However, the best model in the N category, Med42-70B + ZS-COT, significantly outperforms Clinical Big Bird. The Med42-70B + FS, the best model for the M category, has a macro F1 comparable to Clinical-BigBird.

% potential explaination:
% This conforms to findings from the previous study \cite{lehman_we_2023} that small specific models fine-tuned using training data may still outperform LLMs.

\begin{table}[]
\caption{Paired T-test: Difference of Macro F1 between Clinical-BigBird and Best Model selected from Table \ref{tab:performance-comparison-prompting-strategies}}
\label{tab:t-statistics}
\resizebox{\columnwidth}{!}{%
\begin{tabular}{@{}llll@{}}
\toprule
 & \begin{tabular}[c]{@{}l@{}}T Category\\ (Med42-70B + \\ ZS-COT)\end{tabular} & \begin{tabular}[c]{@{}l@{}}N Category\\ (Med42-70B + \\ ZS-COT)\end{tabular} & \begin{tabular}[c]{@{}l@{}}M Category\\ (Med42-70B + \\ FS)\end{tabular} \\ \midrule
Clinical-BigBird & $29.23^{***}$ & $-80.74^{***}$ & -0.70 \\ \bottomrule
\end{tabular}%
}
\end{table}

\subsection{Performance Comparison by Cancer Type}

We conduct further analysis based on the cancer type available in the TCGA dataset. Of the top most frequently diagnosed cancers in U.S. in 2023, breast, prostate, lung and bronchus, and colon and rectum cancer \cite{noauthor_common_nodate}, we select breast (BRCA - breast invasive carcinoma) and lung (LUAD - lung adenocarcinoma) for further analysis as they have the largest number of reports in the dataset. We exclude prostate and colon cancers due to smaller sample sizes and lack of full representation of all T, N, or M categories.  Table~\ref{tab:cancer-specific-performances} shows comparison of the best Med42 prompting strategy vs the benchmark Clinical-BigBird for these cancers. 
By the macro F1 score we note that unlike with the full testing dataset, Med42 outperforms Clinical-BigBird in T classification for both BRCA and LUAD cancers. Consistent with the previous results on the full corpus, Med42  outperforms Clinical-BigBird in N classification. For M classification of BRCA Med42 has better macro F1 than Clinical-BigBird and only slightly worse performance for LUAD. We note that metastatic cases are rare in the TCGA dataset. There are 132 M0 and 6 M1 cases for BRCA and 59 M0 and 4 M1 cases for LUAD. Clinical Big Bird identifies none of the M1 cases in BRCA and LUAD. Med42+FS identifies 4 M1 cases in BRCA, while it cannot identify M1 cases in LUAD. In curating and using this data corpus for TNM classification, Kefeli and Tatonetti~\cite{kefeli_generalizable_2023} report several limitations regarding M01 classification, namely, that many of the pathology reports do not explicitly contain M0 or M1, unlike the case for T and N status, and that TCGA annotations of M01 are occasionally inconsistent with the report text. These factors may explain the low performance of the models in the M classification task.

To understand the variations in performance by type of cancer we hypothesize that reports for a particular type of cancer may have been sourced from the same facility.  However, we are unable to confirm this as TCGA metadata does not contain a facility or source identifier. These results do suggest to us that there is a margin for improvement by customizing the prompting instructions to the type of cancer.  This is a focus of our ongoing work.  

\begin{table*}[t]
\caption{Performance Comparison for two cancer types: BRCA and LUAD}
\label{tab:cancer-specific-performances}
\resizebox{\textwidth}{!}{%
\begin{tabular}{@{}llllllll@{}}
\toprule
 &  & \multicolumn{3}{l}{BRCA} & \multicolumn{3}{l}{LUAD} \\ \midrule
 & Model & Macro P & Macro R & Macro F1 & Macro P & Macro R & Macro F1 \\ \midrule
\multirow{2}{*}{T Category} & Clinical-BigBird & \begin{tabular}[c]{@{}l@{}}0.72\\ {[}0.55,0.90{]}\end{tabular} & \begin{tabular}[c]{@{}l@{}}0.61\\ {[}0.51,0.75{]}\end{tabular} & \begin{tabular}[c]{@{}l@{}}0.64\\ {[}0.53,0.79{]}\end{tabular} & \begin{tabular}[c]{@{}l@{}}0.84\\ {[}0.76,0.91{]}\end{tabular} & \begin{tabular}[c]{@{}l@{}}0.85\\ {[}0.76,0.92{]}\end{tabular} & \begin{tabular}[c]{@{}l@{}}0.84\\ {[}0.75,0.91{]}\end{tabular} \\
 & Med42+ZS-COT & \begin{tabular}[c]{@{}l@{}}0.74\\ {[}0.55,0.89{]}\end{tabular} & \begin{tabular}[c]{@{}l@{}}0.72\\ {[}0.56,0.88{]}\end{tabular} & \begin{tabular}[c]{@{}l@{}}0.72\\ {[}0.56,0.86{]}\end{tabular} & \begin{tabular}[c]{@{}l@{}}0.92\\ {[}0.83,0.97{]}\end{tabular} & \begin{tabular}[c]{@{}l@{}}0.92\\ {[}0.85,0.97{]}\end{tabular} & \begin{tabular}[c]{@{}l@{}}0.91\\ {[}0.83,0.97{]}\end{tabular} \\ \midrule
\multirow{2}{*}{N Category} & Clinical-BigBird & \begin{tabular}[c]{@{}l@{}}0.70\\ {[}0.59,0.79{]}\end{tabular} & \begin{tabular}[c]{@{}l@{}}0.71\\ {[}0.60,0.83{]}\end{tabular} & \begin{tabular}[c]{@{}l@{}}0.69\\ {[}0.58,0.79{]}\end{tabular} & \begin{tabular}[c]{@{}l@{}}0.52\\ {[}0.37,0.80{]}\end{tabular} & \begin{tabular}[c]{@{}l@{}}0.51\\ {[}0.38,0.74{]}\end{tabular} & \begin{tabular}[c]{@{}l@{}}0.50\\ {[}0.38,0.75{]}\end{tabular} \\
 & Med42+ZS-COT & \begin{tabular}[c]{@{}l@{}}0.81\\ {[}0.70,0.91{]}\end{tabular} & \begin{tabular}[c]{@{}l@{}}0.81\\ {[}0.71,0.91{]}\end{tabular} & \begin{tabular}[c]{@{}l@{}}0.80\\ {[}0.70,0.90{]}\end{tabular} & \begin{tabular}[c]{@{}l@{}}0.89\\ {[}0.77,0.98{]}\end{tabular} & \begin{tabular}[c]{@{}l@{}}0.92\\ {[}0.82,0.99{]}\end{tabular} & \begin{tabular}[c]{@{}l@{}}0.89\\ {[}0.79,0.98{]}\end{tabular} \\ \midrule
\multirow{2}{*}{M Category} & Clinical-BigBird & \begin{tabular}[c]{@{}l@{}}0.48\\ {[}0.46,0.49{]}\end{tabular} & \begin{tabular}[c]{@{}l@{}}0.50\\ {[}0.50,0.50{]}\end{tabular} & \begin{tabular}[c]{@{}l@{}}0.49\\ {[}0.48,0.50{]}\end{tabular} & \begin{tabular}[c]{@{}l@{}}0.48\\ {[}0.44,0.49{]}\end{tabular} & \begin{tabular}[c]{@{}l@{}}0.51\\ {[}0.5,0.5{]}\end{tabular} & \begin{tabular}[c]{@{}l@{}}0.51\\ {[}0.40,0.66{]}\end{tabular} \\
 & Med42+FS & \begin{tabular}[c]{@{}l@{}}0.57\\ {[}0.49,0.66{]}\end{tabular} & \begin{tabular}[c]{@{}l@{}}0.74\\ {[}0.45,0.95{]}\end{tabular} & \begin{tabular}[c]{@{}l@{}}0.58\\ {[}0.47,0.71{]}\end{tabular} & \begin{tabular}[c]{@{}l@{}}0.48\\ {[}0.43,0.49{]}\end{tabular} & \begin{tabular}[c]{@{}l@{}}0.51\\ {[}0.5,0.5{]}\end{tabular} & \begin{tabular}[c]{@{}l@{}}0.49\\ {[}0.46,0.50{]}\end{tabular} \\ \bottomrule
\end{tabular}%
}
\end{table*}

\section{Discussion}
In this study we evaluated the ability of open-source LLMs to extract the pathologic TNM stage from real-world pathology reports.  Using one general purpose (Llama-2-70b-chat) and two clinical (Med42-70B and ClinicalCamel-70B) LLMs we compared different standard prompting approaches (zero-shot, few-shots, and zero-shot chain-of-thought) in performing this task.  We also compared the performance of these three LLMs with a pre-trained Clinical-BigBird model that has been fine-tuned with a training set of these reports as a benchmark.

Our findings suggest that open-source generative LLMs are able to perform as well as or even better than the benchmark model that is fine-tuned on the same dataset. The implications of these findings are significant - using locally hosted LLMs, without the need for any training data, we are able to achieve comparable or better performance on extracting TNM stage information from real-world pathology reports.  Because the LLMs we tested are not fine-tuned for this specific task or dataset, we infer that they have the potential to perform well on other clinical notes and tasks, since no training data is needed. 

In comparing prompting approaches, we find that the few-shots approach does not appear to improve the performance of the LLMs significantly.  The TCGA pathology reports come from multiple institutions, and there does not appear to be a standardized structure or formatting system, reflecting the diversity of documentation standards and styles across institutions and pathologists.  We hypothesize because of this diversity, it may be difficult to craft sufficiently generalizable shots for the entire corpus. Verifying and mitigating this issue is the focus of our ongoing work and experiments.

We additionally observed in our experiments that the open-source LLMs appear to be very sensitive to the system instruction and to the prompt structure.  In our experiments, minor variations led to widely varying results, as reported by \cite{sclar_quantifying_2023}.  This has important implications if these LLMs are to be adopted for real world clinical tasks in the healthcare setting. Rigorous prompt testing and engineering needs to be implemented and evaluated to ensure the best performance and outcomes.

\subsection{Strengths and Limitations}
Our study has several strengths. We use real world data that has not been extensively preprocessed to address spelling or formatting issues, and therefore the results from our experiments are likely a true reflection of how these models would perform if embedded in a clinical setting or application. Our approach does not utilize any fine-tuning, thus reducing the human effort required to curate and annotate training data.  We utilize and evaluate locally hosted open-source LLMs, reducing the risks of PHI leakage, and the potential long-term costs involved in using commercial LLMs.

We do note some limitations to our approach.  We evaluate the performance using the testing dataset for a fair comparison with the previous fine-tuned model (i.e., Clinical Big-Bird). Increasing the size of the dataset and evaluating the performance using the whole dataset may allow cancer-specific performance analysis (e.g., Table~\ref{tab:cancer-specific-performances}) for more cancer types, providing insights to develop cancer-specific prompting strategies and instructions. The inference speed using the LLMs is significantly slower than that of the pre-trained Clinical-BigBird. However, developments in techniques to improve inference speed, such as quantization of models and other approaches, may allow us to overcome this barrier.  We only evaluate Llama-based models in our study, and thus the results may not be generalizable to other open-source models.

\section{Conclusion}
Ascertaining cancer stage from pathology reports is a real-world, clinically relevant, and important task in the management of cancer patients, and for researchers who are working to understand and improve cancer outcomes.  The advent of large language models (LLMs) has created an unprecedented opportunity in healthcare for processing of free-text clinical notes and accelerating research.  In this study we demonstrated that locally hosted open-source medical LLMs are able to extract cancer staging information from real world pathology reports without the use of training data. Using standard prompting approaches we obtained comparable performance to a pre-trained model that has been fine-tuned on this same data. The potential to use these open-source medical LLMs across different tasks can be inferred from the high performance obtained without any specific fine-tuning. Exploring how to improve the creation of generalizable few shots to further improve the performance, as well as experimenting with novel and more effective prompting techniques for clinical tasks remains the focus of future work.

\section*{Acknowledgment}

This work was supported in part by the National Science Foundation under the Grants IIS-1741306 and IIS-2235548, and by the Department of Defense under the Grant DoD W91XWH-05-1-023.  This material is based upon work supported by (while serving at) the National Science Foundation.  Any opinions, findings, and conclusions or recommendations expressed in this material are those of the author(s) and do not necessarily reflect the views of the National Science Foundation.

\bibliographystyle{IEEEtran}

\bibliography{references}

\end{document}